\documentclass[pdflatex,sn-mathphys]{sn-jnl}% Math and Physical Sciences Reference Style

\usepackage{lscape}

\jyear{2021}%

\theoremstyle{thmstyleone}%

\theoremstyle{thmstyletwo}%

\theoremstyle{thmstylethree}%

\usepackage{etoolbox}

\begin{document}

\title[ ]{Cross-Sign Language Transfer Learning Using Domain Adaptation with Multi-scale Temporal Alignment}

%%=============================================================%%










%%=============================================================%%

\author{\fnm{Keren} \sur{Artiaga}}\email{kerenartiaga@outlook.com}

\author{\fnm{Yang} \sur{Li}}\email{yangli@sz.tsinghua.edu.cn}

\author{\fnm{Ercan Engin} \sur{Kuruoglu}}\email{kuruoglu@sz.tsinghua.edu.cn}

\author{\fnm{Wai Kin (Victor)} \sur{Chan}}\email{chanw@sz.tsinghua.edu.cn}

\affil{\orgname{Tsinghua-Berkeley Shenzhen Insitute}, \orgaddress{\street{Tsinghua Shenzhen International Graduate School, The University Town, Nanshan District}, \city{Shenzhen}, \postcode{518055}, \country{P.R.China}}}

\miscnote{%
\small
This version of the article has been accepted for publication, after peer review and is subject to Springer Nature's AM terms of use, but is
not the Version of Record and does not reflect post-acceptance improvements,
or any corrections. The Version of Record is available online at:
\url{https://doi.org/10.1007/s11042-023-16703-0}
}

%%==================================%%


%%==================================%%

\abstract{

Sign language serves as a vital means of communication for individuals with hearing impairments, yet recognition resources for the over 100 distinct sign languages are severely lacking. In response, we present our work on sign language recognition using transfer learning and the domain adaptation method TA3N, which utilizes the Temporal Relational Network (TRN) module for aligning multi-scale temporal relations. Our findings highlight the superior performance of Domain Adaptation to neural network-based transfer learning, particularly in improving recognition of American Sign Language (ASL). Our research also identifies the effectiveness of aligning shorter-term temporal features between source and target domains. In addition to using RGB, we conducted experiments using Optical Flow mode for the sign language samples, ultimately determining that RGB outperforms Optical Flow in the majority of cases. Our work aims to improve accessibility and communication for individuals who rely on sign language as their primary mode of communication.}




\maketitle

\section{Introduction}\label{sec1}

The World Health Organization predicts that by 2050, one out of ten people will have some degree of hearing loss. Translation between sign languages and spoken languages is crucial for facilitating communication between sign language users and those who do not use sign language. However, Sign Language Recognition (SLR) is not as extensively researched as other types of action recognition, and SLR studies tend to focus on only a few of the 135 sign languages used worldwide due to the lack of high-resource datasets for most of these sign languages. It is common knowledge that low-resource datasets are prone to overfitting in classification or recognition tasks. Our solution to the low-resource dataset problem for SLR is cross-sign language transfer learning, where we transfer knowledge from higher-resource sign language datasets. This method is domain-specific and differs from state-of-the-art SLR methods that use non-sign language-specific datasets, such as ImageNet. By applying sign language to sign language transfer learning, we can improve SLR by learning high-level, specific visual features that cannot be provided by general large-scale non-sign language datasets, which are limited to low-level, superficial features.

By definition, transfer learning is a machine learning technique that works by applying the knowledge gained in one domain to another. Specifically, this paper aims at achieving effective cross-sign language transfer learning using Domain Adaptation. Our study focuses on Argentine Sign Language (LSA) and Chinese Sign Language (CSL) as the source domains, with American Sign Language (ASL) as the target domain. However, the methodology presented in this paper can be universally applied to any sign language recognition task. We decided on ASL to be the target domain as being a low-resource dataset, it has a less number of samples per class than LSA and CSL, as seen in Table \ref{tab: Sample size video-level}. Domain adaptation is a subset of transfer learning that brings the target task’s distribution closer to that of the source. There are studies \cite{DBLP:journals/corr/abs-1805-06618, cayamcela_lim_2019, nishat_shopon_2020, Bird2020BritishSL} on neural network-based transfer learning, specifically pre-training applied to SLR, and a study on domain adaptation between ASL copy-sign data and ASL native data \cite{9455190}. However, there are currently no studies on the application of domain adaptation between different sign languages.
Moreover, as evidenced by the results of our pre-training experiments \ref{Comparison to other methods}, it is clear that pre-training is not sufficient for domains with different distributions, such as sign languages. Therefore, we deem it advantageous to utilize Domain Adaptation. Our major contribution, specifically, is the use of multi-scale temporal alignment to align the domain of the source sign language to that of the target.

We furthered our domain adaptation with multi-scale temporal alignment work by determining the different effects of aligning the source and target domains’ temporal relations at both short-term and long-term timescales. Temporal relations can be defined as a group of select time-ordered frames of a video. From the results of our experiment, we were able to prove our hypothesis that aligning shorter-term relations instead of longer-term ones is the better approach for improving the accuracy of the target SLR model in all adaptations (from LSA to ASL and from CSL to ASL) and in all input modalities we used in this study.

We experimented in two learning settings: \emph{full-scale transfer learning} and \emph{few-shot transfer learning}. In full-scale transfer learning, we use an 80:20 ratio of target training data to target test data for each k-fold cross-validation. Few-shot transfer learning aims to create an accurate model with a test base using less target training data, mimicking human learning. Compared to full-scale learning, few-shot learning requires less training data and computational resources. We implement few-shot transfer learning with a 20:80 ratio of target training data to target test data.

We used RGB and Optical Flow as input modalities, which can easily be generated from video frames, eliminating the need for additional equipment such as a specialized camera used in several SLR works, such as \cite{Bird2020BritishSL}. We specifically implemented the Gunner Farneback optical flow algorithm\cite{Farnebck2003TwoFrameME}, which has the advantage of being invariant to appearance\cite{SevillaLara2018OnTI}, making generalization easier for low-resource sign language datasets.

The datasets we used in this work are LSA64: A Dataset of Argentinian Sign Language \cite{Ronchetti2016} , Chinese isolated SLR dataset \cite{1, 2, 3} and WLASL300 \cite{li2020word} datasets for LSA, CSL, and ASL, respectively.

To summarize, our contributions are as follows:

(1) We implemented multi-scale temporal alignment (Sec. \ref{Temporal Relations Network}) to conduct domain adaptation between different sign languages

(2) We determined the different effects of aligning temporal relations of the sign languages at both short-term and long-term timescales 

(3) We conducted domain adaptation in two learning settings: \emph{full-scale transfer learning} (Sec. \ref{subsec:Full-scale Domain Adaptation Results}) and \emph{few-shot transfer learning} (Sec. \ref{subsec:Few-shot Domain Adaptation Results})

(4) We used two input modalities for the samples: RGB and Optical Flow (Sec. \ref{Pre-processing of Videos}), and provided a comparison of their performance for each experiment (Sec. \ref{Experiment Results}).

This paper is organized as follows: In Section 2, related works on Sign Language Recognition (SLR) and Domain Adaptation are discussed. Section 3 outlines the technical approach used in the study, including pre-processing of the videos and details of the sign language models. Section 4 presents the dataset, experimentation setup, results, and analysis. Finally, Section 6 provides conclusions and suggestions for future research in this field.

\section{Related Works}

\label{Related Works}

\subsection{Related Works on Sign Language Recognition}

\label{sec:Related Works on Sign Language Recognition}

In SLR, transfer learning is commonly done by pre-training on large-scale datasets like ImageNet of a different domain and using its weights to initialize the target’s weights.

One of the earliest applications of transfer learning in SLR is the creation of ASL word models by Farhadi et al. \cite{farhadi_forsyth_white_2007}  that transfer between different signers as well as different aspects. For these ASL word models, the authors used one subject in the source dataset and one in the target dataset. The subject for the source is an avatar. 

Nishat et al. \cite{nishat_shopon_2020} conducted unsupervised transfer learning to the Bangla sign language dataset (BsDL) to recognize Bangla character sign language such as digits, vowels, and consonants using VGG16 architecture pre-trained on ImageNet. Another work on BsDL is by Das et al. \cite{Das2022AHA} wherein a hybrid approach using deep transfer learning model with random forest classifier is used.

Cayamcela et al. \cite{cayamcela_lim_2019} used a convolutional neural network (CNN) pre-trained on the ImageNet dataset to help translate American Sign Language Alphabet in real time.  Both \cite{cayamcela_lim_2019} and \cite{nishat_shopon_2020} are dealing with static signs where the inputs to the network are still images and not videos.  Recently,  Halvardsson et al. \cite{https://doi.org/10.48550/arxiv.2010.07827}, developed a Swedish Sign Language Alphabet model that uses an InceptionV3 network that is also pre-trained on ImageNet. This study involves 8 subjects and similar to \cite{cayamcela_lim_2019},  \cite{nishat_shopon_2020} and \cite{Das2022AHA}, does not utilize any videos as inputs.

Recent research has shown that pre-training neural networks on large-scale datasets like ImageNet can improve the accuracy of sign language recognition (SLR). This method is currently considered the state-of-the-art in transfer learning for SLR, as demonstrated in \cite{DBLP:journals/corr/abs-1805-06618, https://doi.org/10.48550/arxiv.2010.07827, cayamcela_lim_2019, nishat_shopon_2020, Zakariah2022SignLR, Shania2022TranslatorOI, Thakar2022SignLT, Das2022AHA, Jiang2020FingerspellingIF, Sharma2021IndianSL}. Meanwhile, a study by Suharjito et al. \cite{Suharjito2021SIBISL} uses both ImageNet as well as kinetic dataset - an action recognition dataset, as source for training their Indonesian Signal System recognition model.

However, domain-specific transfer learning is gaining popularity in improving neural network models for SLR. Bird et al. \cite{Bird2020BritishSL} used transfer learning by pre-training with British Sign Language (BSL) to improve recognition of American Sign Language (ASL) using VGG16 architecture. The study included 18-class BSL and ASL datasets, with ASL consisting of one-handed signs from two subjects. The authors found that the late fusion of RGB and Leap Motion approaches (multimodality) in SLR improved their ASL recognition model compared to using RGB or LM alone. They suggested that similarities in movement between the selected BSL and ASL gestures transferred useful knowledge between the two sign languages.

Another domain-specific transfer learning study was conducted by Abdullayeva et al. \cite{GG2022TransferLF}, where they pre-trained their Azerbaijian fingerspelling (alphabet) recognition model on both the ImageNet dataset and an ASL alphabet dataset. To the best of our knowledge, no previous research has explored few-shot transfer learning in SLR. Existing studies are limited by the type of sign language they can recognize (i.e. static signs such as alphabet, characters, and digits), the number of subjects in their datasets, or the type of gestures used. Our work is particularly challenging as we focus on ASL, a low-resource language with two-handed gestures performed by different subjects, with an average of only 1.3 repetitions per sign in the dataset. Additionally, the ASL dataset we used is sourced from various online platforms with different lighting, backgrounds, aspects, and subject distances from the camera. The most relevant work to our study \cite{Bird2020BritishSL} only included one-handed gestures from two signers. See Table \ref{tab: comparisontoothermodel} for a detailed comparison with other studies.

\subsection{Related Works on Domain Adaptation}

\label{Related Works on Domain Adaptation}

Domain Adaptation is a type of Transfer Learning that aims to generalize a model trained in one domain to perform well in another domain. This is necessary because models trained in one domain may not work well in another domain due to domain shift. Domain shift occurs when the source and target domains have different probability distributions, even though they share the same task. To address this issue, domain adaptation adjusts the target data’s distribution to match that of the source data.

While both neural network-based transfer learning and domain adaptation can help networks learn discriminative features for low-resource settings, their execution differs. Neural network-based transfer learning reuses the source task’s features and weights in the target task’s network. Domain Adaptation, on the other hand, attempts to reduce shift between two different domains by re-weighting source features.

Similar to video classification tasks, Sign Language classification can suffer from domain discrepancy in both spatial and temporal aspects. Hence, there is a need for domain adaptation approaches, specifically for video recognition and classification. Currently, there are only a few studies that apply Domain Adaptation to video classification \cite{Sultani2014HumanAR, Xu2016DualMT, Jamal2018DeepDA, Sahoo2021ContrastAM}. Only Temporal Attentive Adversarial Adaptation Network (TA3N) \cite{Chen2019TemporalAA},  Temporal Attentive Moment Alignment Network (TAMAN) \cite{Xu2016DualMT}, and Contrast and Mix \cite{Sahoo2021ContrastAM} are tested on large-scale datasets. TAMAN is created for multi-source video domain adaptation, while Contrast and Mix are designed for unsupervised domain adaptation. Although TAMAN and Contrast and Mix may be applied to a single-source problem, we decided to use TA3N as it incorporates within its architecture an extensible, plug-and-play module called Temporal Relation Network (TRN) \cite{zhou2017temporalrelation} that generates multiscale temporal relations. TRN is based on humans’ ability to connect meaningful transformations in an object or event without observing all changes, known as temporal relational reasoning. TRN aims to discover important temporal relations between frames in videos when applied to neural networks. Technical implementation of TRN is described rigorously in Section \ref{Temporal Relations Network}

Our aim is to determine the optimal timescale (short-term or long-term) for aligning the temporal relations of the source and target domains. To achieve this goal, we concluded that TA3N, which includes the TRN module, is the most suitable architecture for our cross-sign language domain adaptation task.

\section{Technical Approach}

\label{Technical Approach}

\begin{figure}[htbp!]

    \centering

    \includegraphics[scale=0.5]{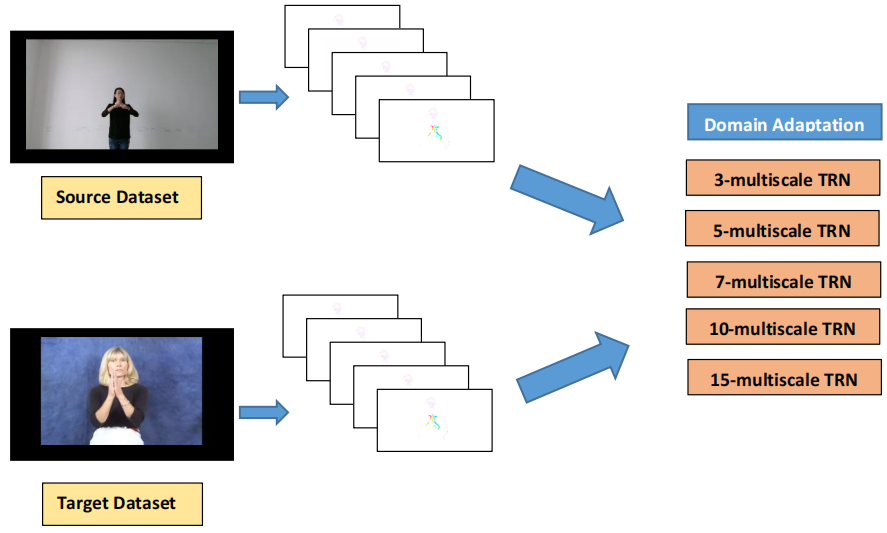}

    \caption{The process involved two steps: pre-processing and domain adaptation, to generate an SLR model. In the pre-processing step, we convert videos into both Optical Flow and RGB frames. The features of these frames are then fed into the Domain Adaptation network five times, each time utilizing a different N-multiscale temporal relations network (TRN) to determine the optimal timescale for aligning the temporal relations of the sign language domains.}

    \label{fig:Two-StepProcess}

\end{figure}

Our technical approach is a two-step process that consists of a pre-processing step and a domain adaptation step as shown in Figure \ref{fig:Two-StepProcess}. In the pre-processing step, we convert videos into RGB frames and Optical Flow frames. After generating the frames, we proceeded with the transfer learning step, where we use the Domain Adaptation approach. Domain Adaptation differs from neural network-based transfer learning in that it is utilized when the source and target tasks share a similar feature space but have distinct probability distributions due to having unrelated datasets. This is particularly relevant in cases where we are adapting the domains of CSL and LSA to ASL. 

We implemented Domain Adaptation in five different multiscale temporal relations: 3, 5, 7, 10, and 15. The shortest term is 3 and the highest term is 15. Section \ref{Temporal Relations Network} describes multiscale temporal relations in detail. The experiment results in Section \ref{Experiment Results} show in detail the effects of N different multi-scale TRN on the recognition accuracy of the target domain.

\subsection{Pre-processing of Videos}

\label{Pre-processing of Videos}

To facilitate cross-sign language transfer learning, we first convert selected individual gesture videos into Optical Flow and RGB frames. To capture the sequential differences between gestures, we experimented with the Optical Flow input mode, which tracks the motion of pixels between frames. We extracted a maximum of 200 frames per sample, as most samples have less than 100 frames.

\begin{figure}[htbp!]

    \centering

    \includegraphics[width=0.75\textwidth]{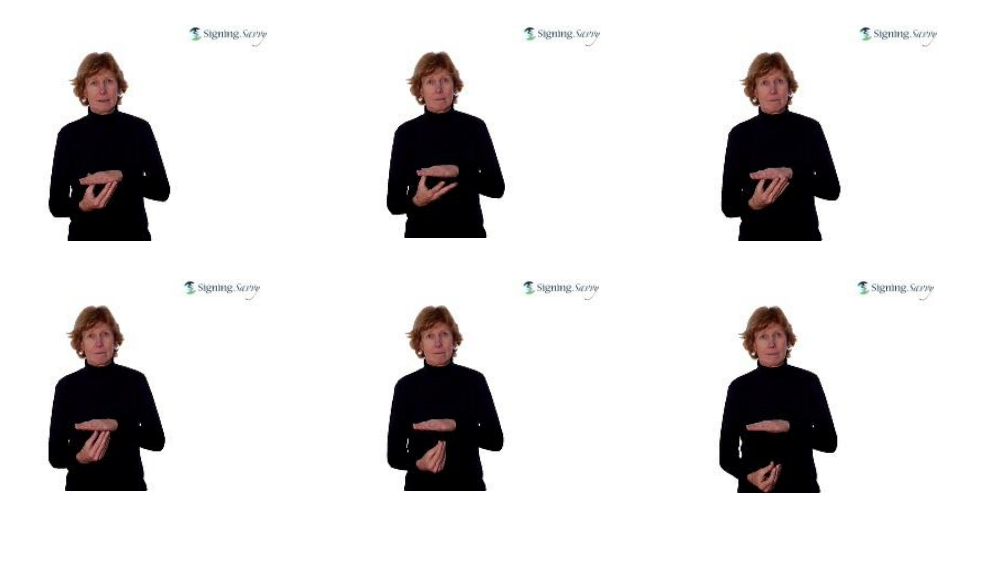}

    \caption{Subsampled RGB frames (starting from left to right)  for the gesture Copy in ASL from WLASL dataset}

    \label{fig:RGB frames Copy in ASL}

\end{figure}

\begin{figure}[htbp!]

    \centering

    \includegraphics[width=0.65\textwidth]{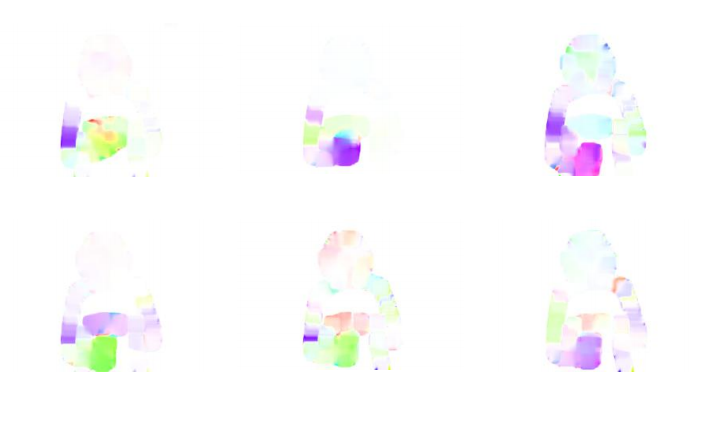}

    \caption{Subsampled optical flow frames (starting from left to right)  for the gesture Copy in ASL from WLASL dataset}

    \label{fig:Optical Flow frames Copy in ASL}

\end{figure}

\begin{figure}[htbp!]

    \centering

    \includegraphics[width=0.8\textwidth]{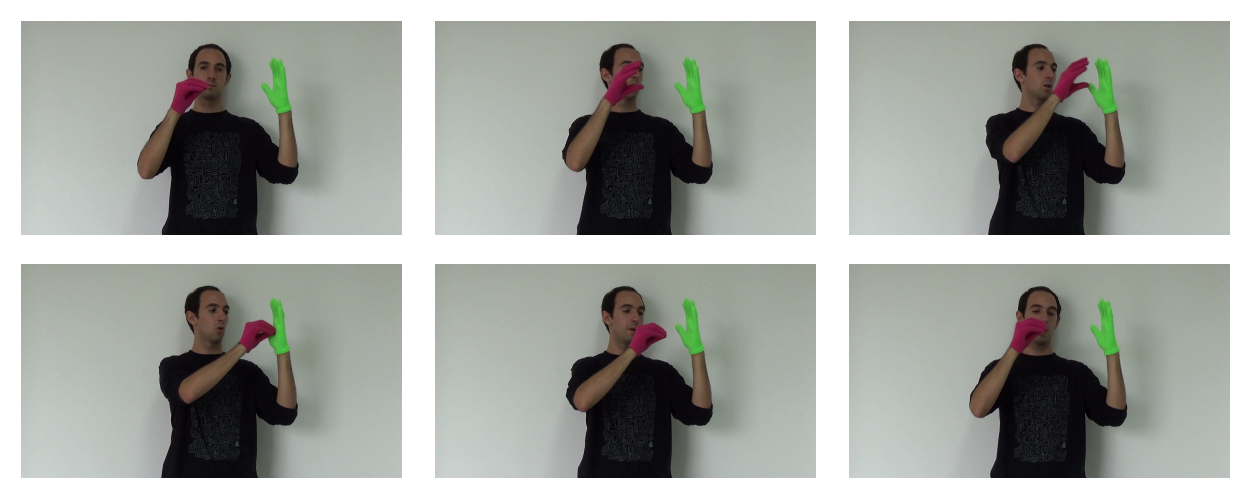}

    \caption{Subsampled RGB frames (starting from left to right)  for the gesture Copy in LSA from LSA64 dataset}

    \label{fig:RGB frames Copy in LSA}

\end{figure}

\begin{figure}[htbp!]

    \centering

    \includegraphics[width=0.8\textwidth]{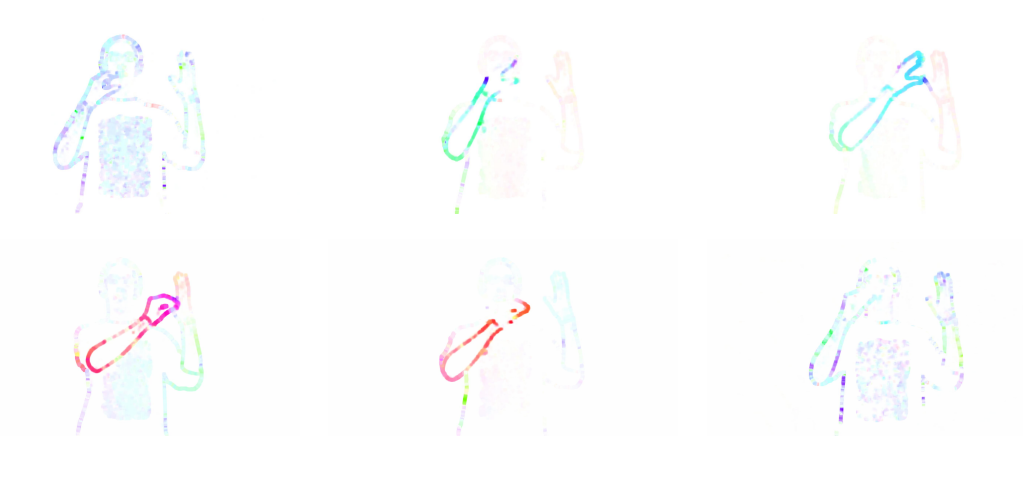}

    \caption{Subsampled optical Flow frames (starting from left to right)  for the gesture Copy in LSA from LSA64 dataset}

    \label{fig:Optical Flow frames Copy in LSA}

\end{figure}

The Figures \ref{fig:RGB frames Copy in ASL}, \ref{fig:Optical Flow frames Copy in ASL}, \ref{fig:RGB frames Copy in LSA} and \ref{fig:Optical Flow frames Copy in LSA} show the subsampled frames from a collection of frames for the gesture Copy in CSL and ASL. For Optical Flow, the backgrounds are changed to white from black for the readers of this paper to see the hand movements more clearly. 

It can be observed that for the optical flow mode of the ASL gesture Copy in Figure \ref{fig:Optical Flow frames Copy in ASL}, the resulting images are pixelated in comparison to its LSA counterpart in Figure \ref{fig:Optical Flow frames Copy in LSA}. The video quality of the samples in ASL is on average poorer than those of CSL and LSA. This is also a reason for selecting ASL as the target domain to improve using Domain Adaptation, in addition to having fewer samples per class.

\subsection{Domain Adaptation for Sign Classification}

\subsubsection{Temporal Relations Network}

\label{Temporal Relations Network}

We are using the Temporal Attentive Adversarial Adaptation Network (TA3N)  \cite{Chen2019TemporalAA} to effectively align two different sign language domains. TA3N aligns multi-scale temporal relation networks of source and target samples by assigning weights to them. The more domain discriminative the temporal relations are, the higher weights they will receive. Section \ref{Detailed TA3N Architecture} explains the implementation of TA3N. The multi-scale temporal relations network used in TA3N is from a separate module called Temporal Relation Network (TRN) \cite{zhou2017temporalrelation} by Zhou et al. 

These temporal relation features are generated by combining the CNN features of N equidistant frames that are sparsely sampled. A single relation feature that represents N number of frames is referred to as an N-frame temporal relation. The function below defines a 2-frame temporal relation:

\[T_{2}(V) = h _{\Phi}(\sum_{i<j}g\Theta (f{i},f{j}))\]
wherein \(T_{2}(V)\) means 2-frame temporal relation of the video \(V\) with n chosen time-ordered frames \(f_{1},f_{2},...,f_{n}\). The function \(g_{\Theta}\) and \(g_{\Phi}\) fuse the features of different time-ordered frames. The uniformly sampled \(i\) and \(j\) frames need not be consecutive so as long as they are sorted chronologically. For a 3-frame temporal relation, below is the corresponding composite function:

\[T_{3}(V) = h _{\Theta}(\sum_{i<j<k}g\Theta (f{i},f{j},f{k}))\]

The final output of the TRN module that is implemented in TA3N is a multiscale TRN which is a fusion of different N-frame temporal relations at multiple time scales. The function below shows the accumulation of N-frame temporal relations at N-time scales:

\[MT_{N}(V) = T_{2}(V) + T_{3}(V )... + T_{N}(V)\]

\begin{figure}[htbp]

    \centering

    \includegraphics[scale=0.60]{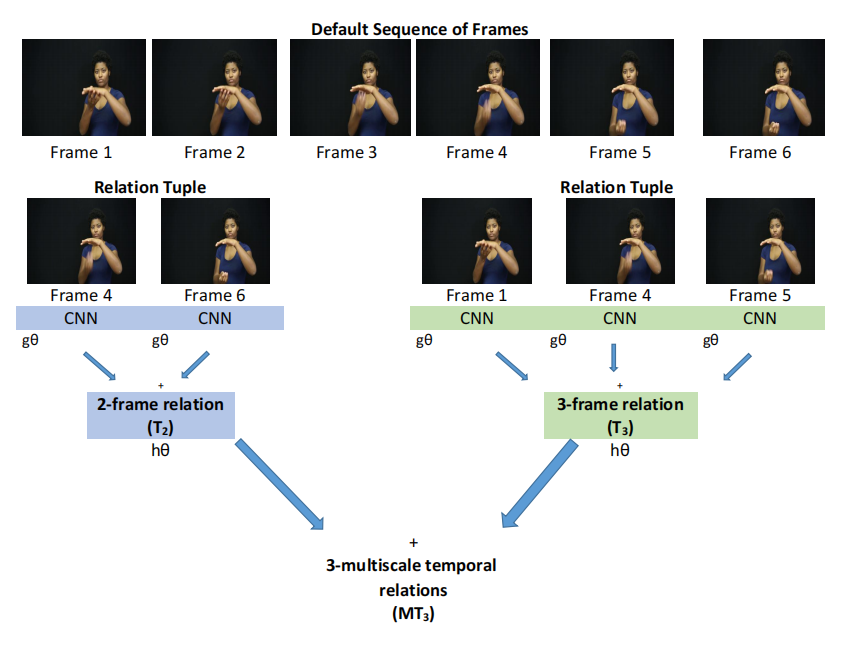}

    \caption{TRN showing the process of generating N-frame temporal relations and finally a multiscale temporal relation for an ASL gesture}

\end{figure}

The TRN module determines the representative frames to be used by converting each frame into features and fusing them into different N-frame temporal relation features, which are then passed into the TRNs. For instance, for 2-frame relations, the TRN module will vote for the combination of two frames that best classify the video.

TA3N focuses on aligning all N-frame temporal relation features \(T_{N}(V)\) of the source domain’s multi-scale TRN \(MT_{N}(V_{s})\) and the target domain’s multi-scale TRN \(MT_{N}(V_{t})\), whereas our objective is to learn the best N of multi-scale TRN \(MT_{N}\) for adapting the domain of one sign language to another.

Further details of the TA3N architecture are described below in Section \ref{Detailed TA3N Architecture}.

\subsubsection{Detailed TA3N Architecture}

\label{Detailed TA3N Architecture}

\begin{figure}[hbt!]

    \centering

    \includegraphics[scale=0.48]{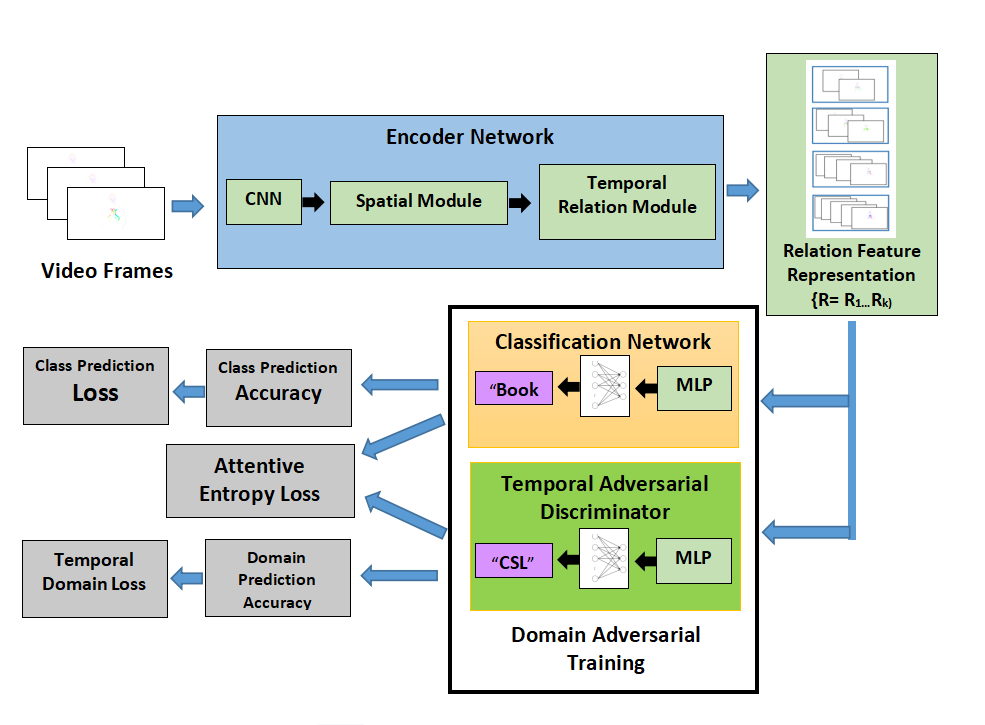}

    \caption{The architecture of TA3N applied on Sign Language Recognition task.}

    \label{fig: TA3N Architecture}

\end{figure}

In Figure \ref{fig: TA3N Architecture}, TA3N applied to our SLR task is shown. The process begins with raw video frames inputted into a convolutional network (CNN) to convert them into frame-level feature vectors, which are then passed to the Spatial Module. This module contains multilayer perceptrons to extract the spatial information of objects, such as hands, in a video, and convert the feature vectors into task-driven feature vectors for video classification.

The task-driven feature vectors are generated and then sent to the Temporal Relation Network (TRN) Module. The TRN module extracts N-frame relations to generate a multiscale TRN \(MT_{N}\), as shown in Figure \ref{fig: TA3N Architecture}. The process of the TRN module is described in detail in Section \ref{Temporal Relations Network}.

The next step involves Domain Adversarial Training for class and domain prediction, where two networks are used: the Temporal Classification Network for class prediction and the Temporal Adversarial Discriminator for domain prediction. Each network consists of a multi-layer perceptron (MLP) with a fully connected layers network that attends to the N-frame temporal relations of the multiscale TRN generated in the previous step.

The network will assign higher weights to N-frame temporal relations that will score low in Temporal Domain Loss which is calculated by the Temporal Adversarial Discriminator. This discriminator will calculate the loss based on the domain prediction. If the domain prediction is correct, the Temporal Adversarial Discriminator will output a low domain loss. Through backpropagation, the Domain Attention block will update the weights of the N-frame temporal relations \(T_{N}(V)\) of the target domain’s multi-scale TRN \(MT_{TN}(V_{t})\) where t means target domain in an attempt to increase the domain loss.

A fully connected layer as seen in Figure \ref{fig: TA3N Architecture} converts the N-frame relations into class predictions to which Prediction Loss is computed. Prediction Loss must be calculated as well because, as discussed in Section \ref{Related Works on Domain Adaptation}, Domain Adaptation works by minimizing the class prediction loss and then maximizing the domain prediction loss.

The final loss to be calculated is called Attentive Entropy Loss which is a product of domain entropy and class entropy. Domain entropy and class entropy are calculated from domain loss and class loss, respectively. TA3N added this type of loss in the architecture to enhance the certainty of videos that have low domain discrepancy to enable it to focus more on videos, specifically their N-frame relations, that have high domain discrepancy by assigning higher weights to them during backpropagation. When the model experiences low domain discrepancies, indicated by lower domain prediction loss and low attentive entropy loss, it implies confusion between the source and target, whereas high domain discrepancies are observed when the model accurately predicts the domain, indicating high discrepancies between the source and target. The network needed to enhance the certainty of low-domain discrepant videos as their N-frame temporal relations do not need attention weights as much as those of high-domain discrepant videos. In Section \ref{Related Works on Domain Adaptation}, we elaborate on why we opted for TA3N for our SLR tasks, as well as its distinguishing features from other related domain adaptation architectures.

\subsubsection{Fine-Tuning TA3N for Sign Language Recognition}

N-multiscale TRN \(MT_{N}(V_{s})\) is based on Temporal Relation Network (TRN) \cite{zhou2017temporalrelation} which is an essential parameter for the TA3N architecture as it is used to determine the number of multiple temporal relations to align between the source and the target domain.  Section \ref{Temporal Relations Network} describes TRN in detail. In the TA3N architecture, the multi-scale TRN parameter is set to 5 by default; therefore, all of TA3N’s experiments with large-scale action recognition datasets are conducted in 5-multiscale TRN. Therefore, we argue that we can fine-tune this value for SLR by using a smaller N-multiscale TRN. For our study, we compared the performance of 3, 5, 7, 10, and 15 multi-scale TRNs to prove our hypothesis that short-term multi-scale TRNs would produce better results for SLR domain adaptation. We refer to 3, 5, and 7 multi-scale TRNs as shorter-term multiscale TRNs while 10 and 15 as longer-term multiscale TRNs. The reason shorter-term multiscale TRNs would perform better is that they would prevent overfitting, especially in the case where sign languages have significant differences from each other. It is worth noting that LSA, CSL, and ASL are from different sign language families. In our experiment results and analysis section ( see Section \ref{Experiment Results} ), we would select the best multi-scale TRN for RGB and Optical Flow in each LSA to ASL and CSL to ASL adaptations.  

Lastly, we implemented TA3N in a supervised manner by setting the \emph{use\_target} parameter to \emph{sV} (i.e. supervised). The default learning implementation of TA3N is unsupervised, that is, the classification loss is calculated solely from the labeled source data. By modifying the \emph{use\_target} parameter to \emph{sV}, the classification loss in our experiments would now be based on the concatenation of the labeled source and target data.

\subsection{Few-Shot Transfer Learning}

We conducted experiments using both full-scale transfer learning and few-shot transfer learning settings. Few-shot transfer learning is a type of Few-Shot Learning (FSL) that aims to achieve good learning performance with a limited training set in a supervised manner \cite{Wang2020GeneralizingFA}. We used few-shot transfer learning to train our SLR model because we wanted it to classify sign language gestures with just a few examples, similar to how humans learn. Our target domain, ASL, has an average of about 14 samples per class. For full-scale transfer learning, we split the target training and test data in an 80/20 ratio. For few-shot transfer learning, we used a 20/80 split, resulting in a limited target training data of approximately 2 samples per class.

\section{Experiments}

\label{Experiments}

TA3N was originally tested on large-scale action recognition datasets, HMDB51 \cite{6126543} and UCF101 \cite{soomro2012ucf101}, using a ResNet-101 architecture. However, its default implementation showed poor performance on sign language recognition. We hypothesize that this is mainly due to the lower number of samples in comparison to HMDB51 and UCF101 datasets, which is a common characteristic among sign language datasets. Hence, we needed to modify the default implementation to fit a more domain-specific, low-resource task of sign language recognition.

We experimented to determine the effects of using shorter-term and longer-term multiscale TRNs on the accuracy of our target sign language recognition model. We also wanted our model to learn from few-shot transfer learning settings. Additionally, we aim to compare RGB inputs against Optical Flow inputs. Section \ref{Experiment Results} will show how our SLR models performed in the following scenarios: (1) short-term versus long-term multiscale TRNs, (2) full-scale versus few-shot transfer learning, (3) RGB vs Optical Flow input mode. 

\subsection{Experiment Setup}

\label{sec:Experiment Setup}

This study utilizes the CSL Isolated Chinese Sign Language Dataset \cite{1, 2, 3} for CSL, the Word-Level American Sign Language (WLASL300) \cite{li2020word} dataset for ASL, and LSA64:A Dataset of Argentinian Sign Language \cite{Ronchetti2016} for LSA. The table presented below displays the classes that we used for the domain adaptation from LSA to ASL and CSL to ASL.

\begin{table}[htbp]

\caption{List of classes used in all transfer learning experiments.}

\label{Tab:List of classes}

\begin{center}

\begin{tabular}{@{}cc@{}}

\toprule

CSL and WLASL300 labels & LSA and WLASL300  labels \\ \midrule

deaf                             & deaf                              \\

red                              & red                               \\

green                            & green                             \\

yellow                           & yellow                            \\

colors                           & colors                            \\

son                              & son                               \\

call                             & call                              \\

milk                             & milk                              \\

none                             & none                              \\

name                             & name                              \\

chair                            & water                             \\

thin                             & man                               \\

white                            & women                             \\

work                             & learn                             \\

blanket                          & country                           \\

coffee                           & where                             \\

friend                           & birthday                          \\

future                           & music                             \\

husband                          & candy                             \\

light                            & catch                             \\

new                              & help                              \\

student                          & dance                             \\

ugly                             & buy                               \\

                                 & copy                              \\

                                 & run                               \\

                                 & give                              \\

\midrule                                 

Total of 23 classes     & Total of 26 classes      \\ \bottomrule

\end{tabular}

\end{center}

\end{table}

To ensure sufficient training and test sets, we used a subset of the larger WLASL dataset, namely WLASL300, which comprises the top 300 classes with the most samples. We selected classes for the experiments based on the labels present in both the source datasets (LSA64 and Isolated Chinese Sign Language) and the target dataset (WLASL300). We identified 26 mutual labels for LSA and ASL datasets, and 23 mutual labels for CSL and ASL datasets.

For the subset of the WLASL300 dataset used in this study, a subject only repeats a gesture on average 1.3 times compared to LSA64 and Chinese isolated SLR, where each subject repeats a gesture 5 times. This demonstrates the generalization ability of the WLASL dataset over other sign language datasets. However, we believe that this ability can cause underfitting, as the average number of samples per class in this dataset is less than 20. In comparison, LSA64 has 50 samples per class, and the Chinese isolated SLR has 250. 

\begin{table}[htbp]

\caption{Training sample sizes and testing sample sizes video-level}

\label{tab: Sample size video-level}

\begin{center}

\begin{tabular}{@{}ccccc@{}}

\toprule

 &

  LSA &

  CSL &

  \begin{tabular}[c]{@{}c@{}}ASL for\\ LSA to ASL\\  DA\end{tabular} &

  \begin{tabular}[c]{@{}c@{}}ASL for \\ CSL to ASL\\ DA\end{tabular} \\ \midrule

Training Sample Size &

  1040 &

  4600 &

  284 &

  257 \\

Testing Sample Size &

  260 &

  1,150 &

  71 &

  64 \\ \bottomrule

\end{tabular}

\footnotetext{DA stands for Domain Adaptation}

\end{center}

\end{table}

Tables \ref{tab: Complexity of the Domain Adaptation Models for LSA to ASL} and \ref{tab: Complexity of the Domain Adaptation Models for CSL to ASL} show the number of learnable parameters in our Domain Adaptation models per N-multiscale TRN. 

\begin{table}[htbp]

\caption{Complexity of the Domain Adaptation Models for LSA to ASL. The number of parameters for each N-multiscale TRN is indicated in the first row, while the number of multiscale TRN for each model is listed in the second row.}

\label{tab: Complexity of the Domain Adaptation Models for LSA to ASL}

\begin{center}

\begin{tabular}{@{}llllll@{}}

\toprule

Number of Parameters & 2582844 & 3895616 & 5732676 & 9471306 & 18323796 \\ \midrule

N-multiscale TRN     & 3       & 5       & 7       & 10      & 15       \\ \bottomrule

\end{tabular}

\end{center}

\end{table}

\begin{table}[htbp]

\caption{Complexity of the Domain Adaptation Models for CSL to ASL. The number of parameters for each N-multiscale TRN is indicated in the first row, while the number of multiscale TRN for each model is listed in the second row.}

\label{tab: Complexity of the Domain Adaptation Models for CSL to ASL}

\begin{center}

\begin{tabular}{@{}llllll@{}}

\toprule

Number of Parameters & 2580534 & 3893306 & 5730366 & 9468996 & 18321486 \\ \midrule

N-multiscale TRN     & 3       & 5       & 7       & 10      & 15       \\ \bottomrule

\end{tabular}

\end{center}

\end{table}

All of our experiments are benchmarked with randomized 5-fold cross-validation, and training time is up to 100 epochs. Our training-to-testing sample size ratio is 80:20 for full-scale domain adaptation, and 20:80 for few-shot domain adaptation. We also have selected a batch size of 20.

For both full-scale and few-shot domain adaptation, we have decided to compare the results from utilizing 3, 5, 7, 10, and 15 multiscale TRNs. Aside from calculating the mean accuracy from each 5-fold cross-validation, we also computed their standard deviations. 

\subsection{Experiment Results and Analysis}

\label{Experiment Results}

For our experiment, we aim to compare the performance between (1) domain adaptation and neural network-based transfer learning, (2)full-scale and few-shot domain adaptation, and (3) N-different multi-scale TRNs. Our analysis will first highlight the best N-multiscale TRN for each group of the experiment: (1) full-scale domain adaptation between LSA to ASL, (2) full-scale domain adaptation between CSL to ASL, (3) few-shot domain adaptation between LSA to ASL, and (4) few-shot domain adaptation between CSL to ASL. Then, in Section \ref{Comparison to other methods}, we will discuss how domain adaptation outperforms neural network-based transfer learning.

We are not only considering improvements from the baseline or non-transfer learning base model in determining the best N-multiscale TRN. We are also identifying the N-multiscale TRN that produced the maximum accuracy for each group of experiments. This approach ensures that improvements over weak baselines or baselines with the lowest classification accuracy are not misleading. The best N-multiscale TRN should produce a domain-adapted model with the highest classification accuracy and a significant improvement over its baseline.

Although the resulting accuracy from our experiments may appear low compared to other studies on Sign Language Recognition (SLR), our transfer learning task is particularly challenging. We experiment with both one-handed and two-handed dynamic word-level signs from the most diverse American Sign Language (ASL) dataset. This diversity includes not only signers or subjects but also video recording conditions such as aspect, lighting, and distance from the camera. These conditions persist because we collect WLASL samples from various sources, including YouTube and educational websites. To our knowledge, no cross-sign language transfer learning study has focused on two-handed dynamic signs. Table \ref{tab: comparisontoothermodel} summarizes the main differences between our target domain ASL and the sign language datasets used in other relevant SLR studies. One critical factor to note is that signers in the WLASL subset we used appeared almost only once per class, with each class containing around 14 video samples. This is in contrast to Bird et al.’s study, the most relevant literature to our work, where their target ASL dataset features only two signers for all classes.

\subsubsection{Full-scale Domain Adaptation Results}
\label{subsec:Full-scale Domain Adaptation Results}

\begin{figure}[htbp!]

    \centering

    \includegraphics[scale=0.4]{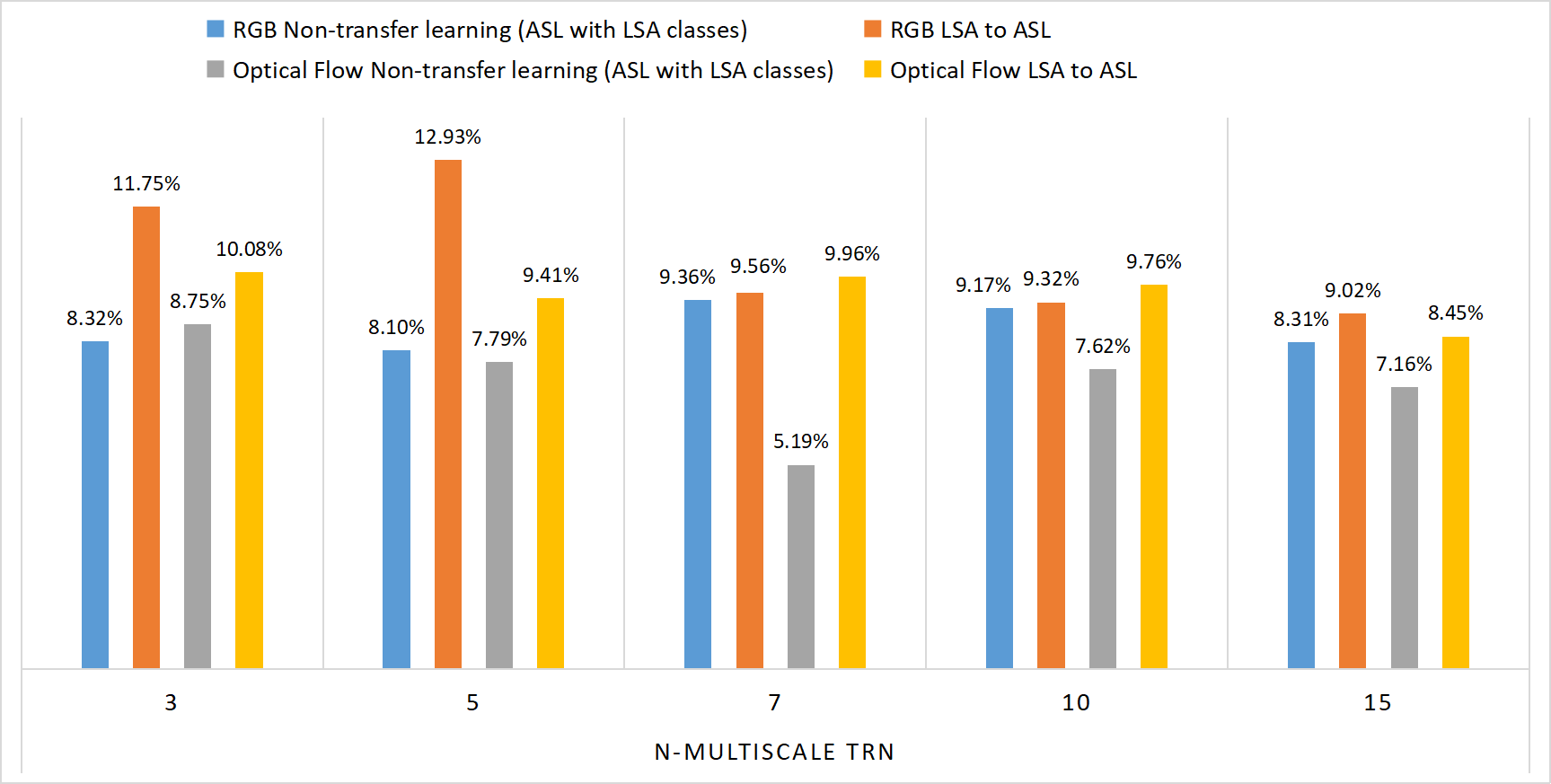}

    \caption{Bar chart showing the Full-scale Domain Adaptation experiment between LSA to ASL across different N-multiscale TRNs}

    \label{fig: FullScaleLSA}

\end{figure}

\begin{figure}[htbp]

    \centering

    \includegraphics[scale=0.4]{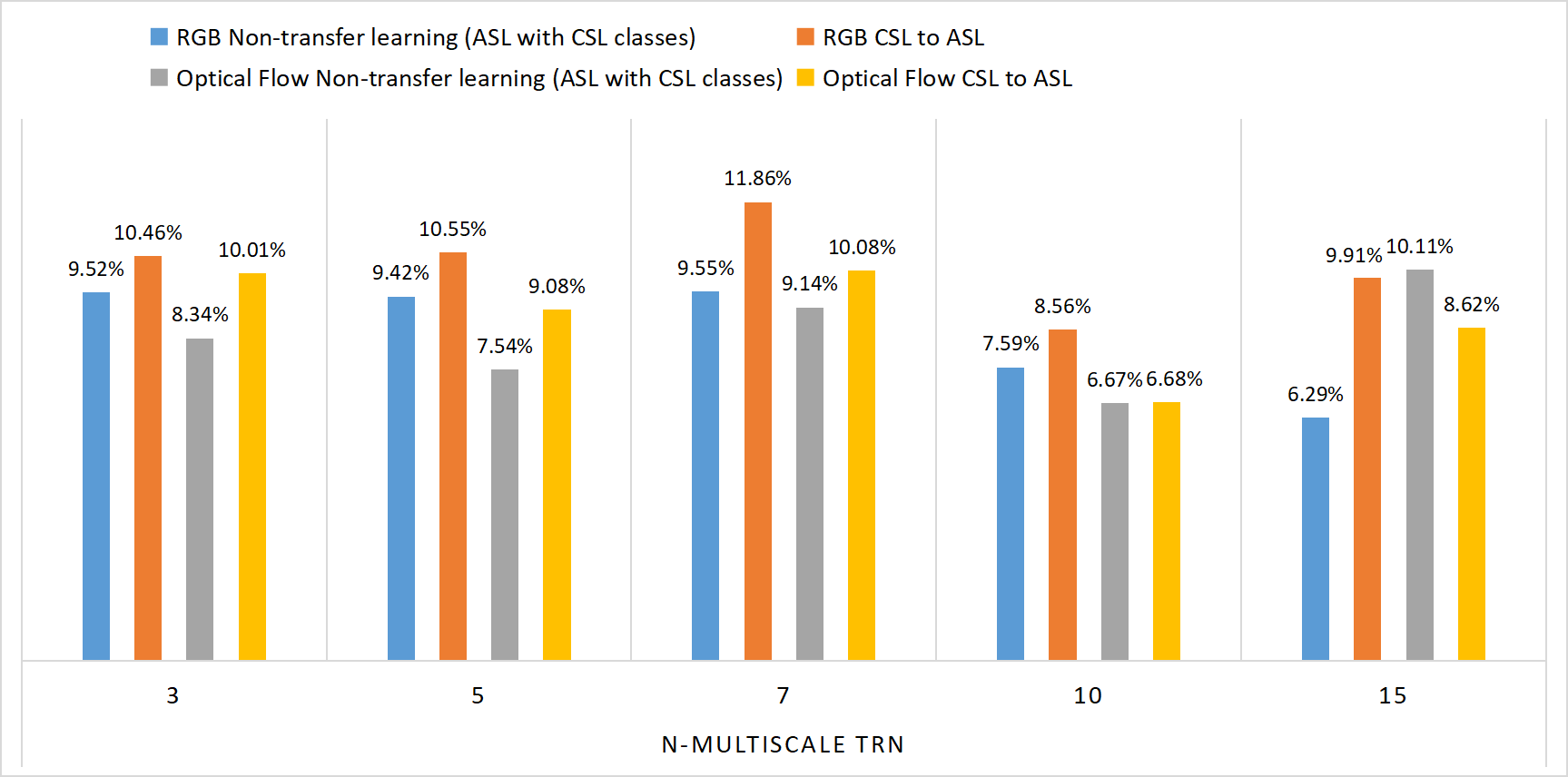}

    \caption{Bar chart showing the Full-scale Domain Adaptation experiment between CSL to ASL across different N-multiscale TRNs}

    \label{fig: FullScaleCSL}

\end{figure}

The bar charts in Figures \ref{fig: FullScaleLSA} and \ref{fig: FullScaleCSL} shows how the classification accuracies of the baseline ASL models and the domain-adapted ASL models change across different N-multiscale TRNs. Conversely, Table \ref{tab: Full-scaleDA} shows all the results of conducting full-scale domain adaptation.

\paragraph{\textbf{Full-scale LSA to ASL domain adaptation}} \medskip

The bar chart in Figure \ref{fig: FullScaleLSA} shows that using the 5-multiscale TRN resulted in the highest classification accuracy of 12.93\% for the domain adapted model in RGB mode, in the LSA to ASL Domain Adaptation group of experiments. This improvement represents a 59.62\% increase over the baseline. Among all RGB domain adaptations for this group, this improvement is the highest, making it easy to conclude that the 5-multiscale TRN is the best N-multiscale TRN for RGB mode in this group.

When analyzing the results of our LSA to ASL full-scale domain adaptation in Optical Flow mode, we found that the 3-multiscale TRN domain adaptation achieved a maximum classification accuracy of 10.08\%, representing a 15.15\% improvement from its baseline. On the other hand, the 7-multiscale TRN domain adaptation showed the highest improvement of approximately 92.017\% from its corresponding baseline. In addition, the 7-multiscale TRN obtained the second-highest classification accuracy of 9.958\%. Given these results, we conclude that the 7-multiscale TRN is the best N-multiscale TRN for LSA to ASL full-scale domain adaptation in Optical Flow mode.

\paragraph{\textbf{Full-scale CSL to ASL domain adaptation}} \medskip
In the bar chart shown in Figure \ref{fig: FullScaleCSL}, it can be observed that the 7-multiscale TRN yielded the highest classification accuracies for CSL to ASL domain adaptation in both RGB (11.86\%) and optical flow mode (10.08\%). Compared to the corresponding RGB baseline, this represents an improvement of 24.19\%, while for optical flow, the improvement is 10.31\%.

The 15-multiscale TRN helped the RGB baseline improve the most at 57.38\% compared to other N-multiscale TRNs. However, the resulting domain-adapted model has the second-lowest accuracy among other domain-adapted models in RGB mode at 9.906\%. Therefore, for RGB mode, 7-multiscale TRN is the best for this group of experiments.

In contrast, for optical flow mode, the 7-multiscale TRN was not the most effective as the 3-multiscale TRN produced an improvement of nearly 20\% from the baseline. Although the 7-multiscale TRN achieved the highest classification accuracy of 10.08\%, the 3-multiscale TRN came in a close second with an accuracy of 10.01\%. Therefore, for this set of experiments, the 3-multiscale TRN is the most effective in domain adaptation for optical flow mode.

It is worth noting that a negative transfer is observed in the optical flow input mode under 15-multiscale TRN, where the domain-adapted model’s accuracy of 8.618\% is lower than its baseline accuracy of 10.112\%.

Overall, shorter-term multiscale TRNs such as 3, 5, and 7, on average, outperform longer-term multiscale TRNs in terms of improvement over strong baselines in Full-scale Domain Adaptation.

\begin{table}[htbp!]

\caption{Overall results of the Full-scale Domain Adaptation experiment.  The second column of the table corresponds to the relevant subsets of ASL, which were selected based on their shared classes with LSA and CSL, as shown in Table \ref{Tab:List of classes}. The third column specifies the type of domain adaptation that was performed where \emph{Non-transfer learning} refers to the baseline.}

\label{tab: Full-scaleDA}

\begin{center}

\resizebox{.95\textwidth}{!}{%

\begin{tabular}{@{}lllllllll@{}}

\toprule

N-multiscale TRN &  &                             &      & 3     & 5     & 7     & 10     & 15    \\ \midrule

\multirow{8}{*}{RGB} &

  \multirow{4}{*}{\begin{tabular}[c]{@{}l@{}}ASL with\\ LSA classes\end{tabular}} &

  \multirow{2}{*}{\begin{tabular}[c]{@{}l@{}}Non-transfer\\ learning\end{tabular}} &

  mean &

  8.32\% &

  8.1\% &

  9.364\% &

  9.174\% &

  8.308\% \\

                 &  &                             & std. & 1.399 & 2.502 & 1.26  & 1.716  & 1.696 \\

                 &  & \multirow{2}{*}{LSA to ASL} & mean & 11.75\% & 12.93\% & 9.564\% & 9.318\%  & 9.024\% \\

                 &  &                             & std. & 1.501 & 1.821 & 1.91  & 2.583  & 2.571 \\

 &

  \multirow{4}{*}{\begin{tabular}[c]{@{}l@{}}ASL with\\ CSL classes\end{tabular}} &

  \multirow{2}{*}{\begin{tabular}[c]{@{}l@{}}Non-transfer\\ learning\end{tabular}} &

  mean &

  9.524\% &

  9.418\% &

  9.55\% &

  7.59\% &

  6.294\% \\

                 &  &                             & std. & 1.64  & 1.261 & 4.922 & 0.396  & 0.5   \\

                 &  & \multirow{2}{*}{CSL to ASL} & mean & 10.46\% & 10.55\% & 11.86\% & 8.56\%   & 9.906\% \\

                 &  &                             & std. & 0.602 & 3.321 & 1.772 & 3.00   & 2.688 \\

\multirow{8}{*}{Optical Flow} &

  \multirow{4}{*}{\begin{tabular}[c]{@{}l@{}}ASL with\\ LSA classes\end{tabular}} &

  \multirow{2}{*}{\begin{tabular}[c]{@{}l@{}}Non-transfer\\ learning\end{tabular}} &

  mean &

  8.754\% &

  7.79\% &

  5.186\% &

  7.618\% &

  7.16\% \\

                 &  &                             & std. & 3.242 & 2.919 & 0.451 & 3.173  & 2.452 \\

                 &  & \multirow{2}{*}{LSA to ASL} & mean & 10.08\% & 9.412\% & 9.958\% & 9.756\%  & 8.45\%  \\

                 &  &                             & std. & 1.33  & 2.45  & 1.99  & 2.129  & 1.123 \\

 &

  \multirow{4}{*}{\begin{tabular}[c]{@{}l@{}}ASL with\\ CSL classes\end{tabular}} &

  \multirow{2}{*}{\begin{tabular}[c]{@{}l@{}}Non-transfer\\ learning\end{tabular}} &

  mean &

  8.342\% &

  7.538\% &

  9.138\% &

  6.672\% &

  10.112\% \\

                 &  &                             & std. & 1.341 & 1.546 & 2.33  & 0.585  & 1.3   \\

                 &  & \multirow{2}{*}{CSL to ASL} & mean & 10.01\% & 9.076\% & 10.08\% & 6.6832\% & 8.618\% \\

                 &  &                             & std. & 1.309 & 2.039 & 1.047 & 1.047  & 1.047 \\ \cmidrule(l){1-9} 

\end{tabular}

}

\footnotetext{Standard deviations correspond to the mean of the 5-fold cross-validation results of each N-multiscale TRN domain adaptation}

\end{center}

\end{table}

\subsubsection{Few-shot Domain Adaptation}
\label{subsec:Few-shot Domain Adaptation Results}

\begin{figure}[!htbp]

    \centering

    \includegraphics[scale=0.40]{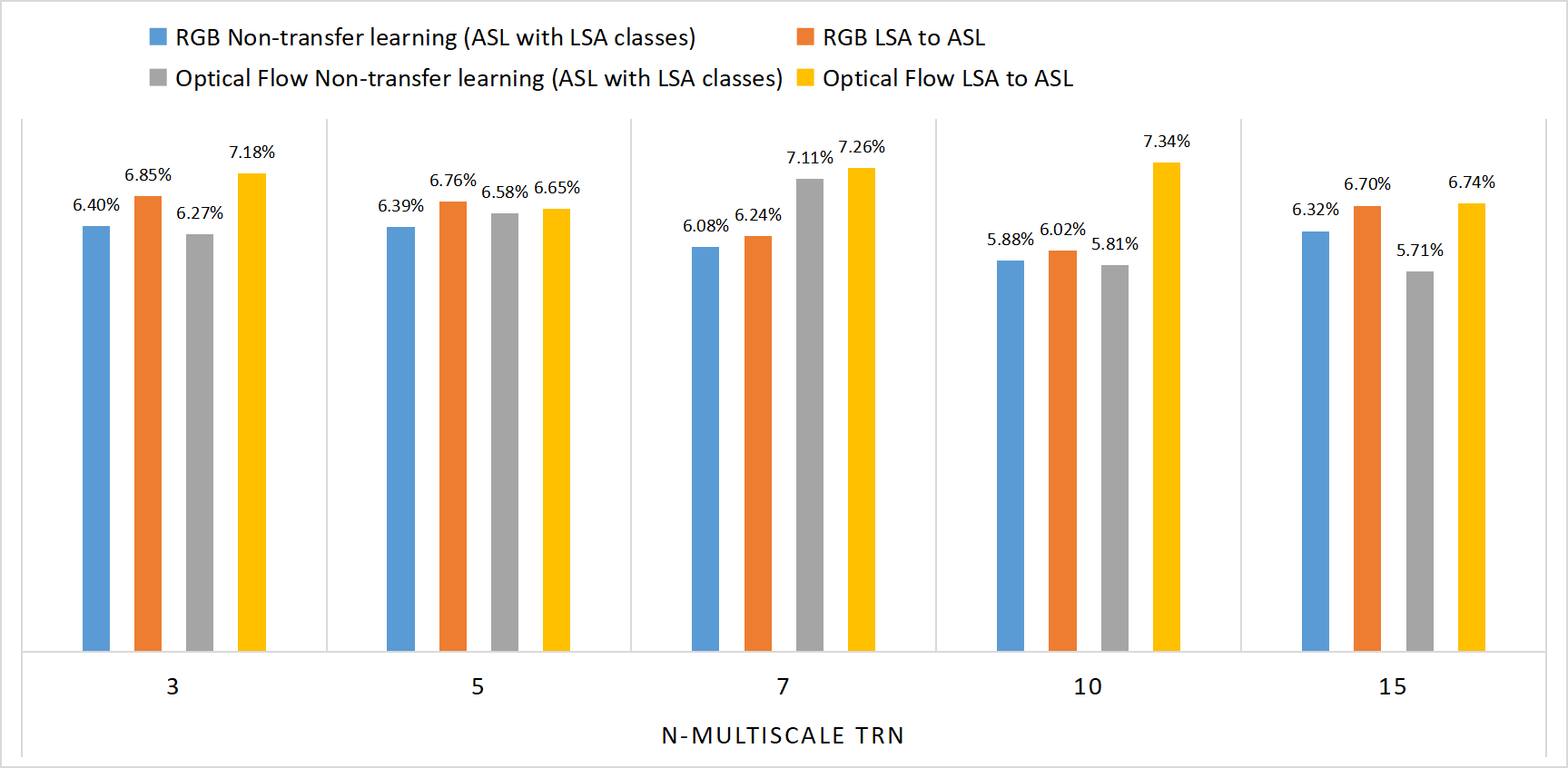}

    \caption{Bar chart showing the Few-shot Domain Adaptation experiment between LSA to ASL across different N-multiscale TRNs}

    \label{fig: FewScaleLSA}

\end{figure}

\begin{figure}[!htbp]

    \centering

    \includegraphics[scale=0.40]{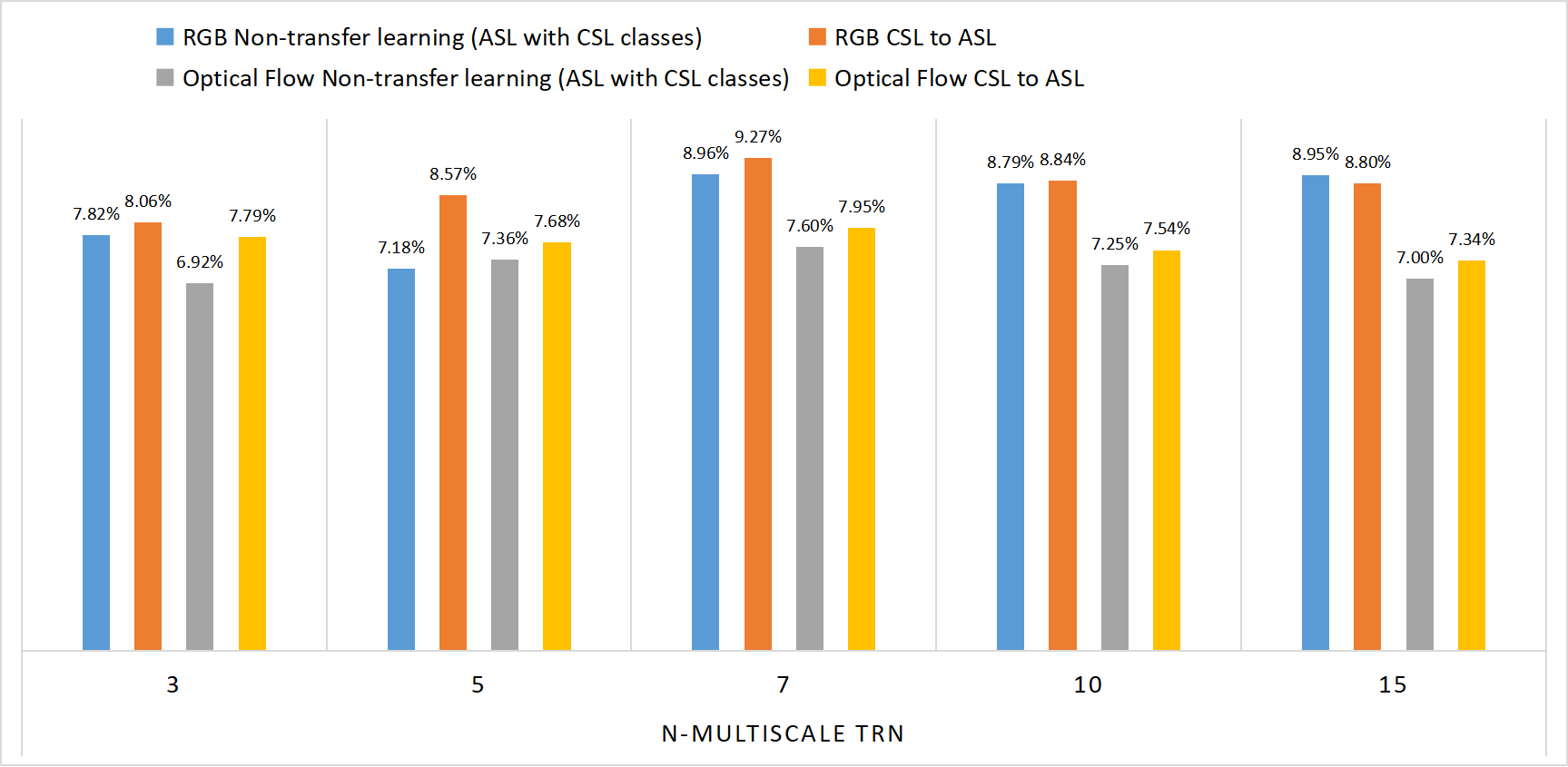}

    \caption{Bar chart showing the Few-shot Domain Adaptation experiment between CSL to ASL across different N-multiscale TRNs}

    \label{fig: FewScaleCSL}

\end{figure}

The bar charts in Figures \ref{fig: FewScaleLSA} and \ref{fig: FewScaleCSL} show how the classification accuracy changes between different N-multiscale TRNs in a few-shot learning setting. Conversely, Table \ref{tab: FewShotDA} shows all the results of our few-scale domain adaptations. 

\paragraph{\textbf{Few-shot LSA to ASL domain adaptation}}\medskip

The bar chart from Figure \ref{fig: FewScaleLSA} shows that for RGB input mode, 3-multiscale TRN achieved the maximum classification accuracy of 6.85\%. Its corresponding baseline is the strongest among all the other comparable baselines, with a classification accuracy of 6.4\%. The percentage difference between this baseline and the domain-adapted model is  7.03\%. From this figure, it is not difficult to conclude that 3-multiscale TRN is the most effective N-multiscale TRN for LSA to ASL few-shot domain adaptation in RGB mode.

Meanwhile, for the Optical Flow mode, 10-multiscale TRN produced the maximum classification accuracy of 7.34\% for the domain-adapted model. This is a 26.48\% increase from the corresponding baseline, which shows the highest improvement in the group. We consider 10-multiscale TRN as the best N-multiscale TRN for this domain adaptation, as it generated the highest classification accuracy as well as the highest improvement over its baseline in this group of experiments.

\paragraph{\textbf{Few-shot CSL to ASL domain adaptation}}
Figure \ref{fig: FewScaleCSL} shows that for CSL to ASL few-shot domain adaptation, the 7-multiscale TRN produced the maximum classification accuracies for all input modes (RGB and Optical Flow). For RGB, the maximum classification accuracy is 9.266\% which is an improvement of 3.41\% from its corresponding baseline, whose accuracy is 8.96\%. For Optical Flow, the maximum classification accuracy achieved is 7.948\%. This accuracy is a 4.55\% increase from the baseline of 7.602\%.

The 5-multiscale TRN yielded the most improvement of approximately 19.32\% from the RGB baseline. However, its corresponding domain-adapted model has the second-lowest classification accuracy among the group at 8.572\%. Therefore, for RGB mode, the 7-multiscale TRN is considered best.

For optical flow mode, we consider the 3-multiscale TRN to be the most effective. The reason is that its domain-adapted model has the second-highest classification accuracy in the group at 7.788\% which translated to an improvement of 12.54\%. Having the second-highest classification accuracy and the highest baseline improvement made 3-multiscale TRN the best for CSL to ASL few-shot domain adaptation.

Similar to full-scale CSL to ASL Domain Adaptation (see Figure \ref{fig: FullScaleCSL}), the 15-multiscale TRN produced domain-adapted models with lower classification accuracy than its baseline (8.796\% vs 8.95\%). Also akin to full-scale domain adaptation, shorter-term multiscale TRNs such as 3, 5, and 7, on average, outperformed longer-term multiscale TRNs in terms of improvement over strong baselines in Few-scale Domain Adaptation as seen in Figures \ref{fig: FullScaleCSL} and \ref{fig: FullScaleLSA}.

\begin{table}[htbp]

\begin{center}

\caption{Overall results of the Few-shot Domain Adaptation experiment. The second column of the table corresponds to the relevant subsets of ASL, which were selected based on their shared classes with LSA and CSL, as shown in Table \ref{Tab:List of classes}. The third column specifies the type of domain adaptation that was performed where \emph{Non-transfer learning} refers to the baseline.}

\label{tab: FewShotDA}

\resizebox{.95\textwidth}{!}{%

\begin{tabular}{@{}lllllllll@{}}

\toprule

N-multiscale TRN &

   &

   &

   &

  3 &

  5 &

  7 &

  10 &

  15 \\ \midrule

\multirow{8}{*}{RGB} &

  \multirow{4}{*}{\begin{tabular}[c]{@{}l@{}}ASL with\\ LSA classes\end{tabular}} &

  \multirow{2}{*}{\begin{tabular}[c]{@{}l@{}}Non-transfer\\ learning\end{tabular}} &

  mean &

  6.4\% &

  6.386\% &

  6.078\% &

  5.88\% &

  6.32\% \\

 &

   &

   &

  std. &

  0.998 &

  0.929 &

  0.77 &

  0.543 &

  0.48 \\

 &

   &

  \multirow{2}{*}{LSA to ASL} &

  mean &

  6.85\% &

  6.764\% &

  6.24\% &

  6.022\% &

  6.702\% \\

 &

   &

   &

  std. &

  0.821 &

  0.686 &

  0.537 &

  0.588 &

  0.267 \\

 &

  \multirow{4}{*}{\begin{tabular}[c]{@{}l@{}}ASL with\\ CSL classes\end{tabular}} &

  \multirow{2}{*}{\begin{tabular}[c]{@{}l@{}}Non-transfer\\ learning\end{tabular}} &

  mean &

  7.816\% &

  7.184\% &

  8.96\% &

  8.792\% &

  8.958\% \\

 &

   &

   &

  std. &

  1.54 &

  1.3 &

  2.234 &

  2.099 &

  1.998 \\

 &

   &

  \multirow{2}{*}{CSL to ASL} &

  mean &

  8.056\% &

  8.572\% &

  9.266\% &

  8.842\% &

  8.796\% \\

 &

   &

   &

  std. &

  0.57 &

  1.245 &

  0.881 &

  0.792 &

  1.584 \\

\multirow{8}{*}{Optical Flow} &

  \multirow{4}{*}{\begin{tabular}[c]{@{}l@{}}ASL with\\ LSA classes\end{tabular}} &

  \multirow{2}{*}{\begin{tabular}[c]{@{}l@{}}Non-transfer\\ learning\end{tabular}} &

  mean &

  6.268\% &

  6.58\% &

  7.108\% &

  5.806\% &

  5.714\% \\

 &

   &

   &

  std. &

  1.132 &

  1.166 &

  0.884 &

  0.906 &

  1.122 \\

 &

   &

  \multirow{2}{*}{LSA to ASL} &

  mean &

  7.184\% &

  6.654\% &

  7.262\% &

  7.344\% &

  6.738\% \\

 &

   &

   &

  std. &

  0.288 &

  1.357 &

  0.236 &

  1.1 &

  0.788 \\

 &

  \multirow{4}{*}{\begin{tabular}[c]{@{}l@{}}ASL with\\ CSL classes\end{tabular}} &

  \multirow{2}{*}{\begin{tabular}[c]{@{}l@{}}Non-transfer\\ learning\end{tabular}} &

  mean &

  6.92\% &

  7.364\% &

  7.602\% &

  7.248\% &

  7.004\% \\

 &

   &

   &

  std. &

  2.688 &

  0.018 &

  1.437 &

  1.999 &

  0.872 \\

 &

   &

  \multirow{2}{*}{CSL to ASL} &

  mean &

  7.788\% &

  7.684\% &

  7.948\% &

  7.536\% &

  7.34\% \\

 &

   &

   &

  std. &

  0.267 &

  0.548 &

  0.601 &

  0.619 &

  0.257 \\ \cmidrule(l){1-9} 

\end{tabular}

}

\footnotetext{Standard deviations correspond to the mean of the 5-fold cross-validation results of each N-multiscale TRN domain adaptation}

\end{center}

\end{table}

Conducting full-scale learning has resulted in domain-adapted models with higher accuracy. Moreover, based on the results of these experiments, it can be concluded that RGB is still superior to optical flow in terms of producing better domain-adapted models. Although LSA and CSL are not related sign languages to ASL, it can be observed that LSA and ASL in general share more similarities in gestures than CSL and ASL. For instance, we discovered that signs such as “Deaf”, “Call”, “Candy”, “Catch”, and “Copy” in our LSA and ASL subsets share notable similarities. Therefore, it is not surprising that LSA as a source helped ASL to reach the highest accuracy in all domain adaptation experiments (see Table \ref{tab: Full-scaleDA}). However, LSA underperformed in our few-shot domain adaptation experiments. The reason is that, since we are only using 20\% of the ASL samples for target training in few-shot learning settings, LSA, being a smaller source domain than CSL, was unable to provide sufficient representations for ASL to learn.

\subsection{Pre-training}

\label{Comparison to other methods}

As part of our study, we also experimented with the pre-training approach wherein the weights learned by a network from the source domain are used as initial weights for training the target domain. This transfer learning method is used by Bird et al. \cite{Bird2020BritishSL}, the first paper to study transfer learning between two sign languages, and other studies that use transfer learning from large-scale datasets such as ImageNet to improve their SLR models \cite{cayamcela_lim_2019, nishat_shopon_2020, DBLP:journals/corr/abs-1805-06618,https://doi.org/10.48550/arxiv.2010.07827}.

To apply this transfer learning technique in our task, we needed to add a Long short-term memory layer on top of the fully-connected layer of the convolutional neural network (CNN). For this method, we utilized a Resnet-50 architecture \cite{He2016DeepRL}. We have decided to use this architecture as it is the most popular deep residual network. 

Similar to our domain adaptation experiment detailed in Section \ref{Experiments}, all of our pre-training experiments are benchmarked with randomized 5-fold cross-validation, and training and testing are continued until there is no more improvement in accuracy after 100 epochs. We also have selected a batch size of 2. For full-scale learning settings, the training-to-test sample size ratio is 80:20 while for few-shot learning settings, the sample size ratio is 20:80.

\begin{table}[htbp]

\caption{5-fold classification accuracies of RGB to RGB and optical flow to optical flow of ASL pre-trained with LSA and CSL as well as the ASL baseline in optical flow (OF) and RGB modes in the full-scale learning setting.}

\label{tab: Full-scale NN-based TL}

\resizebox{.99\textwidth}{!}{%

\begin{tabular}{@{}llllll@{}}

\toprule

 &

   &

  \multicolumn{1}{c}{\begin{tabular}[c]{@{}c@{}}ASL Baseline\\ for LSA to ASL\end{tabular}} &

  \multicolumn{1}{c}{LSA to ASL} &

  \multicolumn{1}{c}{\begin{tabular}[c]{@{}c@{}}ASL Baseline \\ for CSL to ASL\end{tabular}} &

  \multicolumn{1}{c}{CSL to ASL} \\ \midrule

\multirow{2}{*}{RGB} & mean & 6.926\%  & 10.676\% & 9.832\%  & 9.518\%  \\

                     & std. & 2.129  & 7.209  & 4.645  & 4.807  \\

\multirow{2}{*}{OF}  & mean & 10.344\% & 6.094\%  & 15.998\% & 11.748\% \\

                     & std. & 7.409  & 1.601  & 8.425  & 7.489  \\ \cmidrule(l){1-6} 

\end{tabular}

}

\label{tab: Full-scale NN-based TL}

\end{table}

\begin{figure}[htbp]

    \centering

    \includegraphics[scale=0.75]{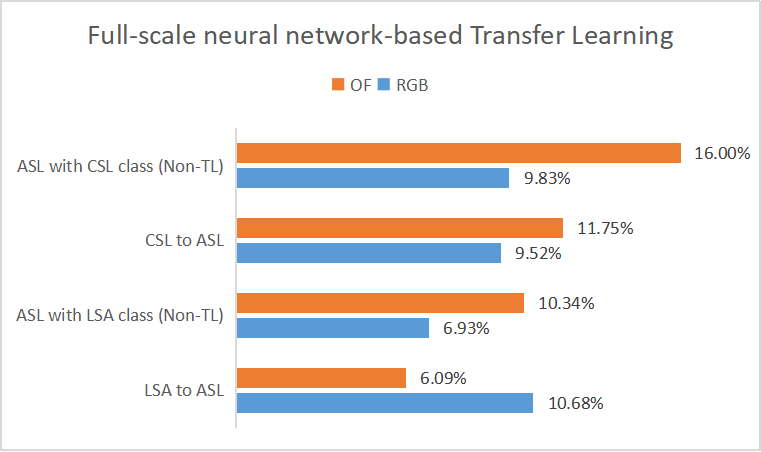}

    \caption{Bar chart showing the accuracies of ASL when pre-trained with LSA and CSL in optical flow (OF) and RGB modes in the full-scale learning setting. Baseline results are also indicated}

    \label{fig: Full-scaleTLgraph}

\end{figure}

\begin{figure}[htbp]

    \centering

    \includegraphics[scale=0.75]{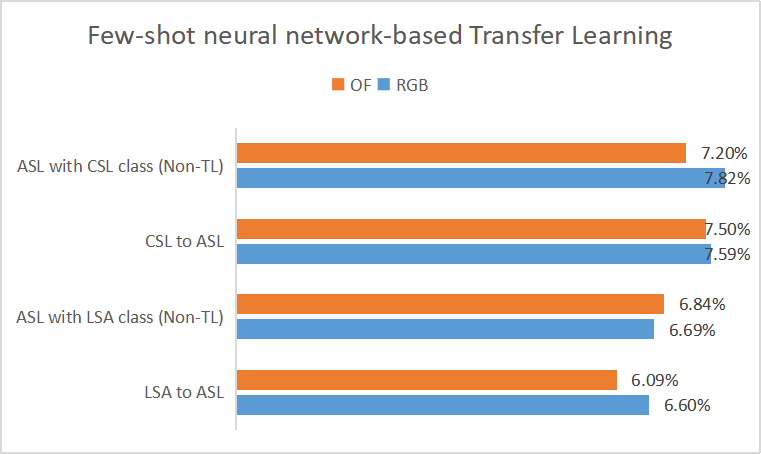}

    \caption{Bar chart showing the accuracies of ASL when pre-trained with LSA and CSL in optical flow (OF) and RGB modes in the few-shot learning setting. Baseline results are also indicated}

    \label{fig: Few-shotTLgraph}

\end{figure}

From Table \ref{tab: Full-scale NN-based TL} and Figure \ref{fig: Full-scaleTLgraph}, we can observe that only ASL pre-trained with LSA in RGB mode produced a positive transfer outcome in full-scale learning settings. On the other hand, for few-shot learning settings, only ASL pre-trained with CSL in optical flow mode resulted in positive transfer as seen in Table \ref{tab: Few-shot NN-based TL} and Figure \ref{fig: Few-shotTLgraph}.

\begin{table}[htbp]

\caption{5-fold classification accuracies of RGB to RGB and optical flow to optical flow of ASL pre-trained with LSA and CSL as well as the ASL baseline in optical flow (OF) and RGB modes in the few-shot learning setting.}

\label{tab: Few-shot NN-based TL}

\resizebox{.99\textwidth}{!}{%

\begin{tabular}{@{}llllll@{}}

\toprule

 &

  \multicolumn{1}{c}{} &

  \multicolumn{1}{c}{\begin{tabular}[c]{@{}c@{}}ASL Baseline\\ for LSA to ASL\end{tabular}} &

  \multicolumn{1}{c}{LSA to ASL} &

  \multicolumn{1}{c}{\begin{tabular}[c]{@{}c@{}}ASL Baseline\\ for CSL to ASL\end{tabular}} &

  \multicolumn{1}{c}{CSL to ASL} \\ \midrule

\multirow{2}{*}{RGB} & mean & 6.686\% & 6.602\% & 7.818\% & 7.59\%  \\

                     & std. & 1.422 & 1.257 & 2.201 & 2.322 \\

\multirow{2}{*}{OF}  & mean & 6.84\%  & 6.086\% & 7.198\% & 7.504\% \\

                     & std. & 1.826 & 1.609 & 1.828 & 2.615 \\ \cmidrule(l){1-6} 

\end{tabular}

}

\end{table}

\begin{table}[htbp]

\caption{Performance matrix of Full-scale Neural network-based Transfer Learning and Domain Adaptation}

\label{tab: Performance matrix Full-scale}

\resizebox{.99\textwidth}{!}{%

\begin{tabular}{@{}lllllll@{}}

\toprule

Tasks                     & \multicolumn{6}{l}{Accuracies}                                  \\ \midrule

Baseline(LSA to ASL) RGB & 6.926  & 8.32     & 8.1      & 9.364    & 9.174     & 8.308     \\

LSA to ASL RGB           & 10.676 & 11.75    & 12.93    & 9.564    & 9.318     & 9.024     \\

Baseline(CSL to ASL) RGB & 9.832  & 9.524    & 9.418    & 9.55     & 7.59      & 6.294     \\

CSL to ASL RGB           & 9.518  & 10.46    & 10.55    & 11.86    & 8.56      & 9.906     \\

Baseline(LSA to ASL) OF  & 10.344 & 8.754    & 7.79     & 5.186    & 7.618     & 7.16      \\

LSA to ASL OF            & 6.094  & 10.08    & 9.412    & 9.958    & 9.756     & 8.45      \\

Baseline(CSL to ASL) OF  & 15.998 & 8.342    & 7.538    & 9.138    & 6.672     & 10.112    \\

CSL to ASL OF            & 11.748 & 10.01    & 9.076    & 10.08    & 6.6832    & 8.618     \\\midrule

                         & NN     & DA 3-TRN & DA 5-TRN & DA 7-TRN & DA 10-TRN & DA 15-TRN \\ \bottomrule

\end{tabular}

}

\footnotetext{NN: Neural Network}

\end{table}

\begin{table}[htbp!]

\caption{Performance matrix of Few-shot Neural network-based Transfer Learning and Domain Adaptation}

\label{tab: Performance matrix Full-scale}

\resizebox{.99\textwidth}{!}{%

\begin{tabular}{@{}lllllll@{}}

\toprule

Tasks                    & \multicolumn{6}{l}{Accuracies}                                 \\ \midrule

Baseline(LSA to ASL) RGB & 6.686 & 6.4      & 6.386    & 6.078    & 5.88      & 6.32      \\

LSA to ASL RGB           & 6.602 & 6.85     & 6.764    & 6.24     & 6.022     & 6.702     \\

Baseline(CSL to ASL) RGB & 7.818 & 7.816    & 7.184    & 8.96     & 8.792     & 8.958     \\

CSL to ASL RGB           & 7.59  & 8.056    & 8.572    & 9.266    & 8.842     & 8.796     \\

Baseline(LSA to ASL) OF  & 6.84  & 6.268    & 6.58     & 7.108    & 5.806     & 5.714     \\

LSA to ASL OF            & 6.086 & 7.184    & 6.654    & 7.262    & 7.344     & 6.738     \\

Baseline(CSL to ASL) OF  & 7.198 & 6.92     & 7.364    & 7.602    & 7.248     & 7.004     \\

CSL to ASL OF            & 7.504 & 7.788    & 7.684    & 7.948    & 7.536     & 7.34      \\\midrule

                         & NN    & DA 3-TRN & DA 5-TRN & DA 7-TRN & DA 10-TRN & DA 15-TRN \\ \bottomrule

\end{tabular}

}

\footnotetext{NN: Neural Network}

\end{table}

Pre-training in both full-scale and few-shot learning settings has resulted in more negative transfers or lower accuracy than positive transfers compared to their respective baselines. In contrast, for domain adaptation, only two instances of 15-multiscale TRN resulted in negative adaptations, while 18 out of 20 N-multiscale TRNs showed improvements in their baselines. Thus, for pre-training, improvements occurred only 25\% of the time, compared to 90\% for domain adaptation.

The standard deviations in this experiment are consistent with those observed in the domain adaptation experiments presented in Section \ref{Experiment Results}, where the test results in the few-shot learning setting were closer to the mean than those in the full-scale learning setting.

\begin{landscape}

\begin{table}[htbp]

\begin{center}

\caption{Comparison of our model to other related Sign Language Recognition models. }\label{tab: comparisontoothermodel}%

\resizebox{1.6\textwidth}{!}{%

\begin{tabular}{@{}llllllllll@{}}

\toprule

Study &

  Source Domain &

  Target Domain &

  Level &

  Dynamic &

  Two-handed &

  Signers &

  \begin{tabular}[c]{@{}l@{}}Number of times\\ a subject repeat\\ a sign\end{tabular} &

  Classes &
  \begin{tabular}[c]{@{}l@{}}Accuracy\\ (Highest Score \\ Achieved) \end{tabular}  \\ \midrule

Bird et al.\cite{Bird2020BritishSL} &

  BSL &

  ASL &

  Word &

  Yes &

  No &

  2 &

  2 &

  18 &
  
  94.44\% \\

Halvardsson et al.\cite{https://doi.org/10.48550/arxiv.2010.07827} &

  ImageNet &

  SSL &

  Alphabet &

  No &

  No &

  5 &

  8 &

  26 &
  85\%   \\

Cayamcela et al.\cite{cayamcela_lim_2019} &

  ImageNet &

  ASL &

  Alphabet &

  No &

  No &

  \begin{tabular}[c]{@{}l@{}}Implied as many.\\ Exact number \\ not indicated\end{tabular} &

  Not indicated &

  26 &
  99.39\% \\

Nishat et al.\cite{nishat_shopon_2020} &

  ImageNet &

  BSDL &

  \begin{tabular}[c]{@{}l@{}}digits, \\ vowel,\\ consonants\end{tabular} &

  No &

  Yes &

  \begin{tabular}[c]{@{}l@{}}Implied as many.\\ Exact number\\ not indicated\end{tabular} &

  Not indicated &

  46 &
  96.57\% \\

Farhadi et al.\cite{farhadi_forsyth_white_2007} &

  ASL Avatar &

  ASL &

  Word &

  Yes &

  Yes &

  1 &

  3 &

  50 & 
  64.17\% \\

Vazquez-Enriquez et al.\cite{Vazquez-Enriquez} &

  ASL &

  \begin{tabular}[c]{@{}l@{}}Turkish and \\ Spanish Sign Language \end{tabular}&

  Word &

  Yes &

  Yes &

  43     &

  Not indicated &

  226 &
  95.24\% (Turkish), 93.91\% (Spanish)\\ 
  
Zakariah et al. \cite{Zakariah2022SignLR} &

Imagenet &

Arabic Sign Language &

Alphabet &

No &

No &

40 &

Not indicated &

32 &

95\% \\

Shania et al. \cite{Shania2022TranslatorOI} &

Imagenet &

Indonesian Sign Language &

Word &

Yes &

Yes &

>2 &

Not Indicated &

11 &

98.5\% \\
 
Abdullayeva et al. \cite{GG2022TransferLF} &
  
 \begin{tabular}[c]{@{}l@{}}Imagenet, \\ ASL letters \end{tabular} &

  Azerbaijan Sign Language &

  Alphabet &

  Yes &

  No &

  Not indicated &

  Not indicated &

  32 &

  88\% \\

Thakar et al. \cite{Thakar2022SignLT} &

Imagenet &

ASL &

Alphabet &

No &

No &

Not indicated &

Not indicated &

29 &

98.7\% \\

Das et al. \cite{Das2022AHA} &

Imagenet &

BsDL &

Digit and characters &

No &

Yes &

Not indicated &

Not indicated &

36 &

91.67\% (character),  97.33\% (digit) \\

Jiang et al. \cite{Jiang2020FingerspellingIF} & 

Imagenet &

CSL &

\begin{tabular}[c]{@{}l@{}}Finger Signs \\ (A to Z, \\ZH, CH, SH, and NG) 
\end{tabular} &

No &

No &

Not indicated &

Not indicated &

30 &

91.48\% \\

Sharma et al. \cite{Sharma2021IndianSL} &

Imagenet &

Indian Sign Language &

Alphabet and digits &

No &

Not indicated &

Not indicated &

Not indicated &

35 &

100\% \\

Suharjito et al. \cite{Suharjito2021SIBISL} &

Imagenet and Kinetic &

Indonesian Signal System &

Not indicated &

Yes &

Not indicated &

2 &

10 &

10 &

97.50\% \\
\midrule

\textbf{Our study} &

  LSA, CSL &

  ASL &

  Word &

  Yes &

  Yes &

  \begin{tabular}[c]{@{}l@{}}109 for WLASL300.\\ For our subsets,\\ on average 11 signers\\ for 14 samples\end{tabular} &

  Average 1.3 &

  23 &
  12.93\% (LSA), 11.86\% (CSL) \\ \bottomrule

\end{tabular}

}

\end{center}

\footnotetext{The number of signers for our study refers to the average number of signers for each class of our subsets. The number of times a signer repeats a sign refers to the number of times a participant recorded a particular sign. In the classes column, we wrote 23 and 26 as we have two sets of classes, one for LSA to ASL transfer and one for CSL to ASL transfer. The average number of samples for each class of our subsets is 14. WLASL300, the dataset we used for our ASL domain, consists of 109 signers and a mean of 17 samples. However, by the time of this study, several of WLASL300’s source links from YouTube and other educational sign language websites are no longer available.}

\end{table}

\end{landscape}

\section{Conclusion}

Our research on transfer learning approaches between different sign languages provided insights on how to improve classification and recognition in low-resource conditions by aligning the source sign language’s spatiotemporal features with the target sign language’s in a multiple-timescale fashion. From our best knowledge, we are the first to study the knowledge adaptability between LSA (Argentine Sign Language) to ASL (American Sign Language) and CSL (Chinese Sign Language) to ASL in both RGB and optical flow modes. We are also the first to conduct a few-shot domain adaptation between sign languages. Applying few-shot learning in SLR is important to advance the field and place it at the levels of other fields such as computational linguistics, natural language processing, and other action recognition areas where many studies are being conducted on the application of few-shot learning.   

The empirical evidence resulting from our study showed that multi-scale temporal Domain Adaptation between LSA and ASL, and CSL and ASL performed significantly better than pre-training ASL with CSL and LSA. The pre-training approach yielded more negative transfers than positive ones. We were also able to prove our hypothesis that aligning shorter-term multiscale TRNs such as 3, 5, or 7 would be more beneficial for the domain adaptation between these sign languages than aligning longer-term multi-scale TRNs such as 10 or 15. This hypothesis holds for all of our experiment settings, such as full-scale and few-shot domain adaptation, and RGB and optical flow input modes (see Section \ref{Experiment Results}). This is a piece of great news for the fields of SLR and action recognition in general as training longer-term multi-scale TRNs consume more GPU resources and time to complete. Moreover, we believe that longer-term multi-scale TRNs will only outperform shorter-term ones in cases where the source and target domains bear more similarities than differences.

Our study also proved our main hypothesis that adapting the domain of one sign language to another - even if they are not from the same sign language family would make a better SLR model than not doing any domain adaptation at all or only pre-training the target domain with another sign language or with a generic large-scale dataset such as ImageNet.

Although our experiment showed that RGB performed better than optical flow (see Section \ref{Experiment Results}), we believe that our study opened more possibilities for the application of optical flow in the field of SLR. In the future, we plan to conduct studies that aim to discover more potential in this type of motion representation.  

Lastly, we envision that our study will be a valuable resource in the real-time translation of sign gestures into words or phrases, or in transcribing videos that feature sign languages. However, to effectively model sign languages, a language modeling technique must be utilized due to differences in grammar structure between spoken languages and sign languages. Nonetheless, these grammar differences can be addressed through domain adaptation. One example is adapting from a sentence-level sign language dataset with spoken language transcription and gloss-level annotation to a dataset with only gloss annotation, to automatically convert gloss-level annotations into transcriptions. Finding similar gloss annotations across datasets is a significant challenge. Alternatively, adapting a non-sign language video with transcriptions to sentence-level sign language videos with gloss-level annotations may be explored in future work.

\backmatter

\section*{Declarations}

\begin{itemize}

\item Conflict of interest/Competing interests

The authors have no relevant financial or non-financial interests to disclose.

\item Funding

This research was funded by the Shenzhen Science and Technology Innovation Commission (JCYJ20210324135011030), Science and Technology Innovation Committee of Shenzhen-Platform and Carrier (International Science and Technology Information Center), High-end Foreign Expert Talent Introduction Plan (G2021032022L), Guangdong Pearl River Plan (2019QN01X890), and National Natural Science Foundation of China (Grant No. 71971127).

\item Ethics approval 

Not applicable

\item Consent to participate

Not applicable

\item Consent for publication

Not applicable

\item Availability of data and materials

All data generated or analyzed during this study are included in these published articles \cite{Ronchetti2016, 1, 2, 3, li2020word} (and its supplementary information files). The subsets we used are detailed in Section \ref{sec:Experiment Setup}. For additional guidance on extracting the subsets from their originating datasets, please contact the authors.

\item Code availability 

The codes used for domain adaptation are based on TA3N \cite{Chen2019TemporalAA}. Our modification includes setting the batch size to 20, the mode of learning to supervised learning, and the value of \emph{num\_segments} to the N-multiscale TRN. The codes for converting videos into RGB and Optical Flow frames are available from this repository, https://doi.org/10.6084/m9.figshare.20223444 . For additional guidance, please contact the authors.
\hfill \break
\hfill \break
All authors contributed to the study's conception and design. Material preparation, data collection, and analysis were performed by Keren Artiaga, Yang Li, Ercan Engin Kuruoglu, and Wai Kin (Victor) Chan. The first draft of the manuscript was written by Keren Artiaga and all authors commented on previous versions of the manuscript. All authors read and approved the final manuscript.

\end{itemize}

%%===================================================%%



%%===================================================%%

%%===========================================================================================%%





%%===========================================================================================%%

\bibliography{sn-bibliography}% common bib file





\end{document}